\documentclass[letterpaper]{article} % DO NOT CHANGE THIS
\usepackage[preprint]{aaai2027}  % DO NOT CHANGE THIS
\usepackage[hyphens]{url}  % DO NOT CHANGE THIS
\usepackage{graphicx} % DO NOT CHANGE THIS
\usepackage{amssymb} % For \blacksquare
\usepackage{amsthm} % For theorem/lemma environments
\usepackage{natbib}  % DO NOT CHANGE THIS AND DO NOT ADD ANY OPTIONS TO IT
\usepackage{caption} % DO NOT CHANGE THIS AND DO NOT ADD ANY OPTIONS TO IT
\usepackage{algorithm}
\usepackage{algorithmic}
\usepackage{newfloat}
\usepackage{listings}
\DeclareCaptionStyle{ruled}{labelfont=normalfont,labelsep=colon,strut=off} % DO NOT CHANGE THIS
\floatstyle{ruled}
\newfloat{listing}{tb}{lst}{}
\floatname{listing}{Listing}
\usepackage{booktabs}
\usepackage{amsmath}
\usepackage{amsfonts}
\usepackage{amsthm}
\newtheorem{theorem}{Theorem}
\newtheorem{proposition}{Proposition}
\newtheorem{assumption}{Assumption}

\title{SDO: Structure-Aware Data Organization for Efficient LLM Post-Training}
\author{
    Jinliang Gao\textsuperscript{\rm 1},
    Ning Yang\textsuperscript{\rm 2}\corresponding,
    Hai Wang\textsuperscript{\rm 3},
    Baili Xiao\textsuperscript{\rm 4},
    Pin Lyu\textsuperscript{\rm 5}\corresponding
}
\affiliations{
    \textsuperscript{\rm 1 2 5}Institute of Automation, Chinese Academy of Sciences\\
    \textsuperscript{\rm 3 4}National University of Defense Technology
}

\begin{document}

\maketitle

\begin{abstract}
Post-training of large language models is expensive, and existing efficiency improvements mainly focus on selecting informative samples or designing training schedules. However, data organization itself is usually treated as a static preprocessing step: embedding-based grouping methods construct fixed partitions before training and cannot adapt to the evolving sample exposure during optimization. As a result, all samples receive similar exposure despite their different optimization needs, leading to redundant updates for some samples while leaving others under-optimized. To address this problem, we propose SDO (Structure-Aware Data Organization), a plug-and-play data organization framework with an exposure-driven feedback mechanism that organizes mini-batch composition and sample exposure according to representation-space structure. SDO operates epoch by epoch on frozen external embeddings, avoiding model warm-up training overhead: within each epoch, locality-aware batching forms coherent mini-batches via KNN neighborhood traversal; across epochs, exposure-balanced scheduling records per-sample participation and reduces the sampling probability of over-exposed samples to preserve long-term coverage. Across SFT, DPO, and GRPO, SDO accelerates convergence, with the largest gains observed in the early-to-mid phase, producing more coherent gradients and more balanced accuracy across question types without permanently excluding training samples.
\end{abstract}


\section{Introduction}

Large language models acquire task abilities through post-training stages including supervised fine-tuning, preference optimization, and reinforcement learning \cite{ouyang2022training, rafailov2023direct, shao2024deepseekmath}, whose cost scales with dataset size and optimization budget \cite{kaplan2020scaling}. Existing data-centric acceleration falls into two lines: data selection, which prioritizes informative samples often via permanent filtering \cite{paul2021deep, xia2024less, chen2024alpagasus, liu2024makes, xie2023data}; and curriculum scheduling, which prescribes a global presentation order based on difficulty or feedback signals \cite{bengio2009curriculum, xu2024contrastive, elhattami2024spaced}. Both focus on selecting what to train and when to present it, but leave the composition of mini-batches within and across epochs unaddressed.

In fact, mini-batch composition directly shapes the optimization signal: mixing unrelated samples produces conflicting gradients, inflating noise and slowing convergence. Recent work has begun to leverage embedding-space locality for sample grouping, demonstrating that organizing batches by representation proximity can improve training efficiency. However, existing methods rely on one-time static partitions computed before training and locked across epochs. This leaves a natural next step open: moving from static ordering to an exposure-driven feedback mechanism that reorganizes the active data pool based on actual sample visitation, and extends across post-training paradigms.

\begin{figure}[t]
\centering
\includegraphics[width=1.0\columnwidth]{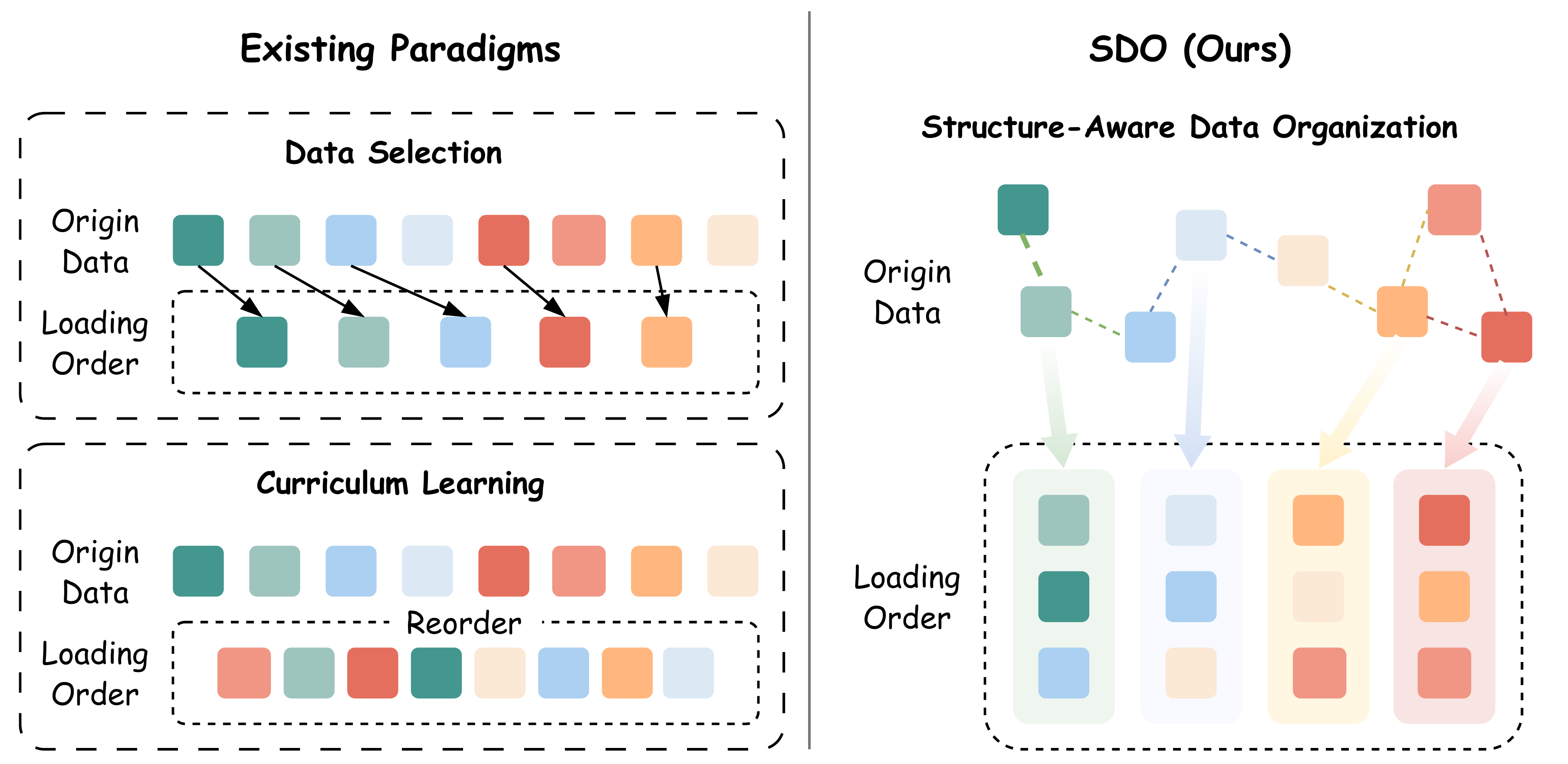}
\caption{Data-centric optimization paradigms vs.\ SDO.}
\label{fig:motivation}
\end{figure}

This paper proposes \textbf{SDO} (\textbf{S}tructure-aware \textbf{D}ata \textbf{O}rganization), a closed-loop data organization framework with an exposure-driven feedback mechanism that organizes data flow at mini-batch and epoch granularities on frozen external embeddings, without altering the learning objective, training schedule, or requiring warm-up (Figure~\ref{fig:motivation}). The main contributions are:

\begin{itemize}

\item \textbf{Revealing the exposure-driven data flow problem.} Existing embedding-based grouping is static and open-loop; exposure-aware pool reconstruction at mini-batch and epoch granularity improves efficiency while preserving coverage, with a theoretical link between representation-space locality and intra-batch gradient consistency.

\item \textbf{A low-cost plug-and-play method with coverage preservation.} SDO combines locality-aware batching via KNN traversal on frozen embeddings and exposure-balanced pool reconstruction across epochs, with coverage preservation. Unlike prior grouping methods, SDO requires no warm-up training and operates on external representations, making it applicable beyond SFT.

\item \textbf{Broad empirical validation across three post-training paradigms.} Across SFT, DPO, and GRPO, SDO accelerates convergence, with consistent gains in the early-to-mid phase, while producing coherent gradients and balanced accuracy across question types; gradient diagnostics confirm improved coherence, and component ablation isolates the contributions of locality and exposure balancing.

\end{itemize}

\section{Related Work}

\subsection{Data-Centric Post-training Optimization}

Data selection and coreset construction improve post-training efficiency by prioritizing informative samples. Early importance sampling methods showed that assigning sampling probabilities according to sample contribution can reduce stochastic gradient noise \cite{katharopoulos2018not}. Recent LLM-oriented approaches estimate sample utility via model-based quality assessment, gradient influence, uncertainty, or distribution matching; representative examples include AlpaGasus \cite{chen2024alpagasus}, LESS \cite{xia2024less}, DEITA \cite{liu2024makes}, DSIR \cite{xie2023data}, and DELIFT \cite{agarwal2025delift}. These methods achieve promising data efficiency by selecting or reweighting samples, often at the cost of permanently discarding a portion of the training set.

Curriculum learning and data scheduling instead regulate the temporal order of presentation. With LLMs, recent work designs schedules according to preference difficulty, reasoning complexity, or model feedback \cite{xu2024contrastive, elhattami2024spaced, feng2024citing, croitoru2026curriculum}, demonstrating that the temporal organization of data can affect optimization outcomes. \cite{dai2026demystifying} studies global sequence ordering (e.g., stair-case or saw-tooth schedules from pre-computed scores) for LLM pre-training and SFT, while \cite{gao2025prompt} selects intermediate-difficulty prompts for RL post-training via a learned value model; both operate at the sequence or prompt level without mini-batch composition or exposure balancing. \cite{tang2026dataorg} analyzes organizational schedules and balanced sampling in multimodal instruction tuning, revealing capability trade-offs that motivate coverage-preserving designs.

Both lines, however, focus on what to train and when, leaving how samples are dynamically grouped into mini-batches and recycled across epochs unaddressed. A broader data-centric survey \cite{luo2025survey} confirms that exposure-aware, batch-level organization remains an underexplored dimension in the data-centric landscape. Orthogonally, recent work improves the post-training objective itself, refining the preference loss formulation~\cite{azar2023ipo, ethayarajh2024kto}, introducing token-level decompositions~\cite{zeng2024tokenlevel, yang2026tokenimportance}, or modifying RL advantage estimation and clipping~\cite{yu2025dapo, zhang2025gspo}; SDO is complementary to these advances, as it modulates data organization rather than loss formulation and can be combined with any of them.

\subsection{Representation Structure for Mini-batch Organization}

The connection between mini-batch composition and optimization noise has long been studied in stochastic optimization.  Lower gradient noise yields more consistent trajectories and faster convergence~\cite{johnson2013accelerating, schmidt2017minimizing}, motivating sample-level weighting, selection, and global controls such as batch size and learning rate~\cite{smith2018don}. These approaches treat samples independently, overlooking the structural relationships that directly shape the coherence of each update.

Representation spaces encode semantic neighborhoods~\cite{khandelwal2020generalization, sorscher2022beyond, coleman2020selection} useful for batch construction. Cluster-GCN~\cite{chiang2019clustergcn} forms dense subgraph mini-batches for GNNs, while EP-Order~\cite{ye2026eporder} clusters samples via HDBSCAN for SFT improvement; however, both rely on fixed partitions and target limited settings. SDO extends this locality insight into an exposure-driven, feedback-based framework spanning SFT, DPO, and GRPO without warm-up or permanent filtering.

\begin{figure*}[t]
\centering
\includegraphics[width=1.0\textwidth]{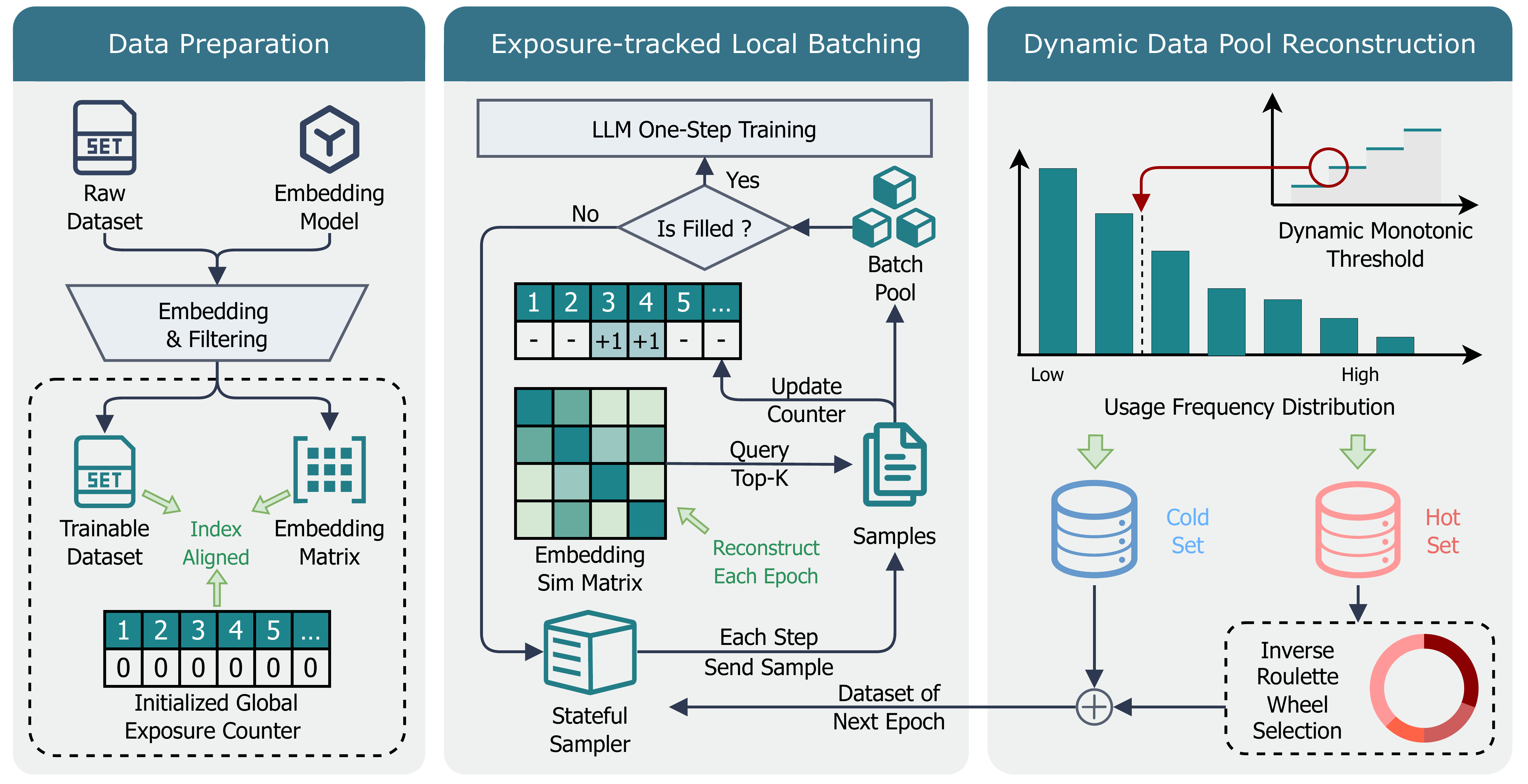}
\caption{
Overview of SDO. The framework consists of three stages:
(1) \textit{Data Preparation}, where samples are mapped into a fixed representation space and initialized with a global exposure tracker;
(2) \textit{Exposure-Tracked Local Batching}, where semantically neighboring samples are organized into coherent mini-batches while recording their historical optimization exposure;
and (3) \textit{Exposure-Driven Pool Reconstruction}, where the training pool is restructured according to accumulated exposure statistics to maintain long-term data diversity.
}
\label{fig:workflow}
\end{figure*}

\section{Method}
\label{sec:method}

SDO implements a closed-loop data organization framework operating at two granularities on frozen external representations: within epochs, it forms semantically coherent mini-batches via KNN-based neighborhood traversal; across epochs, it rebalances sample participation through an exposure ledger that records actual visitation and temporarily down-weights over-explored regions. It is noteworthy that the encoder $f_{\mathrm{enc}}(\cdot)$ is a fixed, pre-trained sentence embedding model computed only once before training, and the closed-loop property refers to data flow where historical sample exposure is fed back to reconstruct the active data pool. Let the training set be $\mathcal{D}_0=\{x_i\}_{i=1}^{N}$, where each sample $x_i=(p_i, r_i)$ consists of a prompt $p_i$ and a response $r_i$. Only the prompt $p_i$ is encoded into a frozen representation $\mathbf{z}_i=f_{\mathrm{enc}}(p_i)\in\mathbb{R}^d$, with $d$ the embedding dimension and $\mathbf{E}=[\mathbf{z}_1,\dots,\mathbf{z}_N]^\top$ the embedding matrix. SDO maintains an active data pool $\mathcal{D}^{(e)}$ and a cumulative exposure ledger $U^{(e)}=\{u_i^{(e)}\}_{i=1}^{N}$ at each epoch $e$, and performs an exposure-driven data flow transformation $\mathcal{T}:(\mathcal{D}^{(e)}, U^{(e)})\rightarrow(\mathcal{D}^{(e+1)}, U^{(e+1)} )$. SDO operates before, during, and after each epoch, as shown in Figure~\ref{fig:workflow}.

\subsection{Topology-aware Neighborhood Construction}
\label{subsec:neighborhood}

At the beginning of each epoch, a local neighborhood structure is built over the current pool. For each sample $x_i$, its $K$ nearest neighbors among all other samples $x_j$ in $\mathcal{D}^{(e)}$ are identified under cosine similarity,
\begin{equation}
\mathcal{N}_K(i)
=
\operatorname*{arg\,topK}_{j\neq i}
\left(
\frac{\mathbf{z}_i^\top \mathbf{z}_j}
{\|\mathbf{z}_i\|_2 \|\mathbf{z}_j\|_2}
\right),
\label{eq:knn}
\end{equation}
where $K$ is the neighborhood size and $j$ indexes candidate samples in $\mathcal{D}^{(e)}$. The resulting set $\mathcal{N}_K(i)$ provides a lightweight approximation of the local topology in representation space and serves as the structural foundation of the subsequent data flow.

\subsection{Exposure-tracked Local Batching}
\label{subsec:batching}

During the epoch, SDO traverses the neighborhood graph to build coherent mini-batches while tracking per-sample participation. It sequentially selects anchor samples and aggregates their neighborhoods $\mathcal{N}_K(i)$ into a batch buffer $\mathcal{B}$. Each anchor is activated once per epoch via a visitation mask $\mathbf{m}\in\{0,1\}^N$. Overlapping neighborhoods create smooth transitions between adjacent batches, as boundary samples may contribute to multiple steps within the same epoch. Once the buffer reaches size $B$, a mini-batch $\mathcal{B}_t$ is extracted and fed into the underlying objective. SDO maintains a global exposure ledger recording cumulative participation,

\begin{equation}
u_i^{(e)} = \sum_{t=1}^{T_e} \mathbb{I}(x_i\in\mathcal{B}_t),
\label{eq:exposure}
\end{equation}

where $T_e$ is the number of steps in epoch $e$ and $\mathbb{I}(\cdot)$ is the indicator function. $U^{(e)}$ forms a long-term memory of historical data flow for the inter-epoch regulation below.

\subsection{Dynamic Data Pool Reconstruction}
\label{subsec:reconstruction}

After the epoch, the accumulated exposure is used to prevent locality-aware batching from over-concentrating on frequently visited regions. Specifically, SDO reconstructs the active pool at the end of each epoch using the statistics $U^{(e)}$. With an epoch-dependent threshold $\tau_e=(e+1)\Delta\tau$ that rises monotonically across epochs, where $\Delta\tau$ is a fixed exposure increment controlling how quickly the cold/hot boundary rises, samples are split into a cold set and a hot set over the full dataset $\mathcal{D}_0$ using the global ledger:
\begin{equation}
\begin{aligned}
\mathcal{D}_{\mathrm{cold}}^{(e)}&=\{x_i\in\mathcal{D}_0:u_i^{(e)}<\tau_e\},\\
\mathcal{D}_{\mathrm{hot}}^{(e)}&=\{x_i\in\mathcal{D}_0:u_i^{(e)}\geq \tau_e\}.
\end{aligned}
\label{coldhotset}
\end{equation}
The cold set is fully retained to preserve under-explored regions. For the hot set, SDO performs temporary inverse-exposure resampling with probability
\begin{equation}
p_i^{(e)}=
\frac{1/u_i^{(e)}}
{\sum_{j\in\mathcal{D}_{\mathrm{hot}}^{(e)}}1/u_j^{(e)}},
\label{eq:resampling}
\end{equation}
so that samples with higher historical exposure receive lower retention probability for the next epoch. A fixed proportion $r\in(0,1]$ of $\mathcal{D}_{\mathrm{hot}}^{(e)}$ is drawn without replacement and combined with the cold set to form the next pool,
\begin{equation}
\mathcal{D}^{(e+1)}
=
\mathcal{D}_{\mathrm{cold}}^{(e)}
\cup
\operatorname{Sample}(\mathcal{D}_{\mathrm{hot}}^{(e)},p_i^{(e)},r),
\label{eq:reconstruction}
\end{equation}
which completes the transformation $\mathcal{T}$. Critically, this is a temporary frequency adjustment rather than permanent filtering. The global ledger $U^{(e)}$ tracks cumulative exposure for all $N$ samples in $\mathcal{D}_0$, including those not in the current pool whose $u_i$ remains frozen while excluded. Since the threshold $\tau_e=(e+1)\Delta\tau$ rises monotonically, it eventually exceeds the frozen $u_i$ of any excluded sample, at which point the sample re-enters via the cold set. Algorithm~\ref{alg:sdo} summarizes the procedure, where $\theta$ denotes model parameters and $T$ the total training epochs.

\begin{algorithm}[t]
\caption{SDO: Structure-aware Data Organization}
\label{alg:sdo}

\textbf{Input}: Dataset $\mathcal{D}_0$, frozen embeddings $\mathbf{E}$, epochs $T$, mini-batch size $B$, neighborhood size $K$, threshold increment $\Delta\tau$, retention ratio $r$ \\
\textbf{Output}: Trained parameters $\theta$

\begin{algorithmic}[1]
\STATE Initialize global exposure $\mathbf{u}\leftarrow \mathbf{0}$; pool $\mathcal{D}^{(0)} \leftarrow \mathcal{D}_0$
\FOR{$e=0$ to $T-1$}
    \STATE Build KNN neighborhoods $\mathcal{N}_K(i)$ for all $x_i\in \mathcal{D}^{(e)}$ from $\mathbf{E}$
    \STATE Initialize visitation mask $\mathbf{m}\leftarrow \mathbf{0}$, batch buffer $\mathcal{B}\leftarrow \emptyset$
    \FOR{each $x_i\in \mathcal{D}^{(e)}$ with $m_i=0$}
        \STATE Mark $m_i\leftarrow1$; append $\{x_i\}\cup\mathcal{N}_K(i)$ to $\mathcal{B}$ \COMMENT{anchor included; neighborhoods overlap intentionally}
        \WHILE{$|\mathcal{B}| \geq B$}
            \STATE Extract mini-batch $\mathcal{B}_t$ of size $B$ from $\mathcal{B}$
            \STATE Update exposure: $u_j\leftarrow u_j+1$ for all $x_j\in\mathcal{B}_t$
            \STATE Optimize: $\theta\leftarrow\mathrm{Optimizer}(\theta,\mathcal{B}_t)$; remove $\mathcal{B}_t$ from $\mathcal{B}$
        \ENDWHILE
    \ENDFOR
    \STATE Update threshold $\tau_e \leftarrow (e+1)\Delta\tau$
    \STATE Classify all $x_i\in\mathcal{D}_0$ by global exposure: $\mathcal{D}_{\mathrm{cold}}^{(e)}=\{x_i\mid u_i<\tau_e\}$, $\mathcal{D}_{\mathrm{hot}}^{(e)}=\{x_i\mid u_i\geq\tau_e\}$
    \STATE Compute $p_i \propto 1/u_i$ for $x_i\in\mathcal{D}_{\mathrm{hot}}^{(e)}$; sample $\mathcal{D}_{\mathrm{hot}}^{\prime}$ of size $\lceil r|\mathcal{D}_{\mathrm{hot}}^{(e)}| \rceil$ without replacement
    \STATE Reconstruct pool: $\mathcal{D}^{(e+1)}=\mathcal{D}_{\mathrm{cold}}^{(e)}\cup\mathcal{D}_{\mathrm{hot}}^{\prime}$
\ENDFOR
\STATE \textbf{return} $\theta$
\end{algorithmic}
\end{algorithm}

\subsection{Theoretical Justification}

The analysis targets why locality-aware batching improves optimization. The central intuition is that mixing samples from different semantic regions produces conflicting gradients that partially cancel, reducing the effective update norm. SDO groups samples by representation proximity, which should mitigate this cancellation.

\subsubsection{Data Mixture Model and Gradient Conflict}

Let $F(\theta)=\frac{1}{N}\sum_{i=1}^{N}\ell_i(\theta)$ be the training objective over $N$ samples, with per-sample gradient $g_i(\theta)=\nabla_\theta\ell_i(\theta)$. To formalize the effect of mini-batch composition, assume the data can be partitioned into $G$ latent semantic clusters $C_1,\dots,C_G$, where cluster $g$ contains $N_g$ samples with proportion $\pi_g=N_g/N$. Define the cluster-wise average gradient and variance:
\begin{equation}
\begin{aligned}
& \mu_g(\theta)=\frac{1}{N_g}\sum_{i\in C_g}g_i(\theta),
\\ & \sigma_g^2=\frac{1}{N_g}\sum_{i\in C_g}\|g_i(\theta)-\mu_g(\theta)\|_2^2.
\label{eq:cluster_stats}
\end{aligned}
\end{equation}
The full-data gradient is $\nabla F(\theta)=\sum_{g=1}^{G}\pi_g\mu_g(\theta)$. The cluster dispersion quantifies the degree to which cluster gradients diverge:
\begin{equation}
\begin{aligned}
\Phi(\theta)&=\sum_{g=1}^{G}\pi_g\|\mu_g(\theta)\|_2^2-\|\nabla F(\theta)\|_2^2
\\&=\sum_{g<h}\pi_g\pi_h\|\mu_g(\theta)-\mu_h(\theta)\|_2^2.
\end{aligned}
\label{eq:dispersion}
\end{equation}
Large $\Phi(\theta)$ indicates strong inter-cluster gradient conflict. The cluster partition is only an analytical abstraction; SDO operates on the KNN graph in $\mathbf{z}$-space.

To link representation-space organization to gradient-space behavior, we assume local gradient smoothness:

\begin{assumption}[Local Gradient Smoothness]\label{ass:smoothness}
$\|g_i(\theta)-g_j(\theta)\|_2 \leq L\|\mathbf{z}_i-\mathbf{z}_j\|_2$ for any sample pair, where $\mathbf{z}_i$ is the frozen embedding of $x_i$ and $L$ is a Lipschitz constant.
\end{assumption}

Assumption~\ref{ass:smoothness} is a sufficient condition that links representation-space locality to gradient similarity, enabling the theoretical analysis below. In practice, the validity of this assumption depends on the encoder's ability to capture task-relevant semantics; the gradient diagnostics in the Gradient Behavior Analysis section provide empirical support for the embedding-gradient association it posits. Under this assumption, an SDO mini-batch with small representation diameter has bounded gradient variation (Lemma~1, stated and proved in the Technical Supplement).

To quantify the optimization consequence, define the \textit{intra-batch gradient conflict} of a mini-batch $B$:
\begin{equation}
\Gamma(B)=\frac{1}{|B|(|B|-1)}\sum_{i\neq j\in B}\|g_i(\theta)-g_j(\theta)\|_2^2.
\label{eq:conflict}
\end{equation}
The following identity links conflict to gradient magnitude (proof in the Technical Supplement):
\begin{equation}
\|\hat{g}\|_2^2 = \bar{a} - \frac{|B|-1}{2|B|}\,\Gamma(B),
\label{eq:identity}
\end{equation}
where $\hat{g}=\frac{1}{|B|}\sum_{i\in B}g_i$ and $\bar{a}=\frac{1}{|B|}\sum_{i\in B}\|g_i\|_2^2$. Larger $\Gamma(B)$ reduces the batch gradient norm, the key mechanism through which batch composition affects optimization.

\subsubsection{Conflict Reduction via Locality-Aware Batching}

\begin{theorem}[Conflict Reduction and Gradient Magnitude under Idealized Cluster-Pure Batching]\label{thm:conflict}
Under the mixture model and Assumption~\ref{ass:smoothness}, consider an idealized locality-aware batching strategy that draws each mini-batch $B_{\mathrm{ideal}}$ from a single cluster. Let $B_{\mathrm{rand}}$ denote mini-batches drawn by uniform random shuffling, both of size $B$ and from the same data pool, with $\hat{g}_{\mathrm{rand}}$ and $\hat{g}_{\mathrm{ideal}}$ their corresponding batch gradient estimators. Then:
\begin{align}
&\mathbb{E}\big[\Gamma(B_{\mathrm{ideal}})\big] = 2\sum_{g=1}^{G}\pi_g\sigma_g^2, \label{eq:conflict_ideal}\\
&\mathbb{E}\big[\Gamma(B_{\mathrm{rand}})\big] = 2\sum_{g=1}^{G}\pi_g\sigma_g^2 + 2\,\Phi(\theta), \label{eq:conflict_rand}\\
&\mathbb{E}\big[\|\hat{g}_{\mathrm{ideal}}\|_2^2\big] - \mathbb{E}\big[\|\hat{g}_{\mathrm{rand}}\|_2^2\big]
= \frac{|B|-1}{|B|}\,\Phi(\theta) \;\geq\; 0. \label{eq:magnitude_gap}
\end{align}
\end{theorem}

Theorem~\ref{thm:conflict} characterizes the theoretical limit of conflict reduction under ideal cluster-pure batching: the expected intra-batch gradient conflict is reduced by $2\Phi(\theta)$ and the gradient norm increases by $\frac{|B|-1}{|B|}\Phi(\theta)$ relative to random shuffling. In practice, SDO's KNN-based batching approximates this ideal, with the approximation quality depending on the encoder's ability to map semantically similar samples to nearby regions and the resulting batch purity. The trade-off between neighborhood size and exploration diversity is analyzed in the ablation study below.

\subsubsection{Exposure-Balanced Coverage}

The ideal-case analysis of Theorem~\ref{thm:conflict} provides a plausible mechanism for the acceleration effect (a larger gradient norm does not by itself guarantee faster convergence), but also reveals a risk: repeated visits to the same clusters may cause overfitting to dominant patterns. The exposure-balanced scheduling addresses this.

\begin{proposition}[Exposure Monotonicity and Coverage Preservation]\label{prop:exposure}
Let $u_i^{(e)}$ be the cumulative exposure of $x_i$ at the end of epoch $e$, and $p_i^{(e)}$ its retention probability in the hot-set resampling. Then: (i) $p_i^{(e)}$ is monotonically decreasing in $u_i^{(e)}$; and (ii) every sample with $u_i^{(e)}<\tau_e$ is fully retained in $\mathcal{D}^{(e+1)}$, so no sample is permanently excluded.
\end{proposition}

Proposition~\ref{prop:exposure} shows that over-exposed samples receive lower retention probability while under-exposed samples are preserved, preventing long-term concentration.

Full proofs of all theoretical results, additional discussions (KNN approximation gap, connection to SGD convergence, aggregation vs.\ diversification), and gradient collection details are provided in the Technical Supplement.

\textbf{Scalability.} For large-scale post-training, the KNN graph can be constructed using an approximate nearest neighbor (ANN) index (e.g., FAISS with HNSW or IVF-PQ) built offline for each data pool, reducing construction complexity from $O(N^2)$ to $O(N\log N)$ while maintaining neighbor recall and bringing the overhead to a negligible level.

\section{Experiments}
\label{sec:experiments}

\subsection{Experimental Setup}
\label{sec:exp_setup}

\begin{table*}[t]
\centering
\footnotesize
\caption{Convergence comparison across three paradigms (mean$\pm$std). ``Gain'': final improvement of SDO over baseline.}
\label{tab:main}
\setlength{\tabcolsep}{3pt}
\begin{tabular*}{\textwidth}{@{\extracolsep{\fill}}llccccc@{}}
\toprule
Dataset / Metric & Method & Early & Mid & Late & Final & Gain \\
\midrule
GSM8K & GRPO & $72.30{\pm}1.36$ & $76.24{\pm}1.04$ & $81.05{\pm}0.66$ & $82.41{\pm}0.20$ & --- \\
Acc.\,$\uparrow$ & GRPO+SDO & $73.08{\pm}0.47$ & $78.09{\pm}1.52$ & $82.00{\pm}0.29$ & $82.89{\pm}0.50$ & \textbf{+0.48} \\
\midrule
UltraFeedback & DPO & $0.028{\pm}0.004$ & $0.150{\pm}0.009$ & $0.264{\pm}0.008$ & $0.314{\pm}0.012$ & --- \\
Margin\,$\uparrow$ & DPO+SDO & $0.033{\pm}0.006$ & $0.167{\pm}0.008$ & $0.307{\pm}0.012$ & $0.358{\pm}0.017$ & \textbf{+0.044} \\
\midrule
UltraFeedback & SFT & $1.526{\pm}0.011$ & $1.180{\pm}0.004$ & $1.121{\pm}0.001$ & $1.095{\pm}0.000$ & --- \\
Loss\,$\downarrow$ & SFT+SDO & $1.503{\pm}0.003$ & $1.175{\pm}0.003$ & $1.123{\pm}0.001$ & $1.096{\pm}0.001$ & \textbf{$\approx$0} \\
\bottomrule
\end{tabular*}
\end{table*}

\textbf{Datasets, model, and encoder.}
Evaluation covers three post-training paradigms. GRPO uses GSM8K \cite{cobbe2021gsm8k} (6{,}796 training examples); accuracy is reported on a fixed 500-example validation split sampled from the test partition, while per-cluster balance is evaluated on the full 1{,}319-example test set. DPO and SFT use 6{,}000 preference pairs from UltraFeedback \cite{cui2024ultrafeedback} (5{,}908 after filtering prompts exceeding 512 tokens); reward margin ($r_{\text{chosen}}-r_{\text{rejected}}$ averaged over the test set) and test loss are evaluated on the test set. All experiments use Qwen3.5-4B. Prompt embeddings are generated offline once by zembed-1-embedding (2560-dim, $\ell_2$-normalized) from prompt text only, and kept frozen. For DPO, each chosen--rejected pair remains intact within the same mini-batch (details in the Technical Supplement).

\textbf{Baselines and protocol.}
The baseline uses uniform shuffling in an otherwise identical pipeline (same model, dataset, hyperparameters, seed). Both methods are evaluated at matched checkpoints. SDO introduces three parameters: neighborhood size $K$, exposure threshold increment $\Delta\tau$, and hot-set retention ratio $r$, with defaults $K{=}4$, $\Delta\tau{=}2$, $r{=}0.2$. All main experiments use this default without task-specific tuning, demonstrating the plug-and-play nature of SDO. For diagnostic analyses focusing on exposure redistribution and gradient behavior in the Gradient Behavior Analysis and Ablation Studies sections, we use $r{=}0.1$, which provides a more balanced exposure distribution and clearer visualization of the underlying mechanism. All runs use a single RTX 4090 (48\,GB); per-setting hyperparameters (batch size, LR, steps) are in the Technical Supplement. Main results (Table~\ref{tab:main}) are mean$\pm$std over three seeds $\{419, 617, 917\}$; per-cluster balance and ablation analyses use a single representative seed (617).

\subsection{Training Efficiency across Post-training Paradigms}
\label{sec:training_efficiency}

\begin{figure*}[t]
\centering
\includegraphics[width=1.0\textwidth]{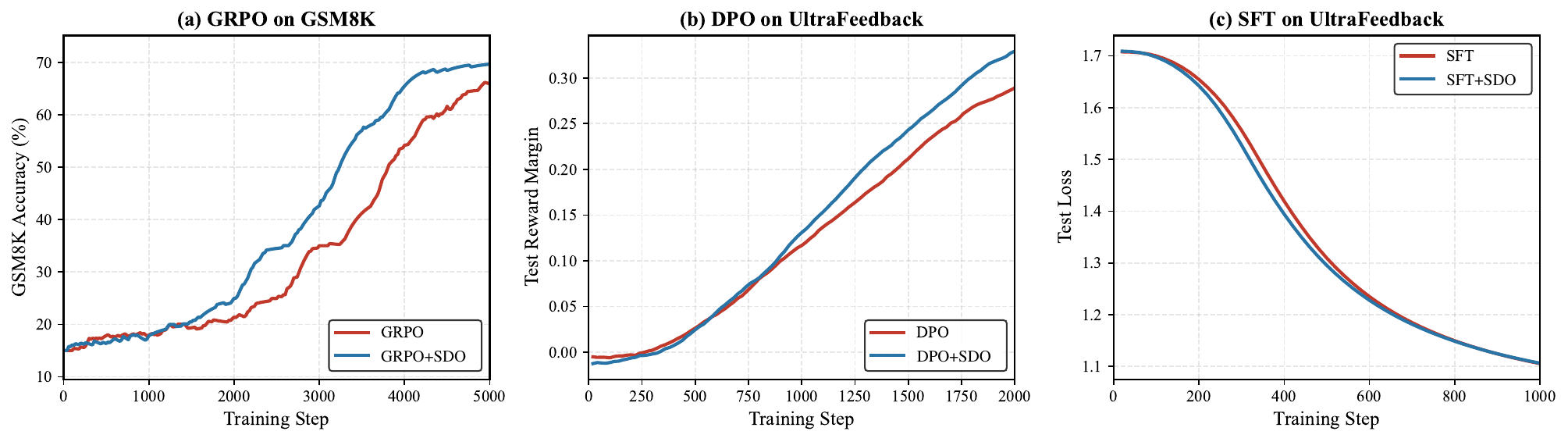}
\caption{Convergence curves across three post-training paradigms. GRPO/GSM8K reports validation accuracy, while DPO and SFT report test reward margin and loss, respectively. SDO consistently converges faster than the baseline in all three settings. (a) GRPO/GSM8K. (b) DPO/UltraFeedback. (c) SFT/UltraFeedback.}
\label{fig:efficiency}
\end{figure*}

The first question is whether SDO alone accelerates post-training. Figure~\ref{fig:efficiency} and Table~\ref{tab:main} report convergence curves and checkpoint values across three settings.

\textbf{GRPO.} SDO outperforms the baseline at every checkpoint, with the largest gap at 3k steps ($+1.85$) and a final gain of $+0.48$. Figure~\ref{fig:efficiency}(a) shows that SDO reaches the baseline accuracy roughly 20\% earlier, indicating faster convergence. Its full run takes 1799 min versus 1786 min for the baseline, adding only 0.7\% runtime.

\textbf{DPO.} SDO produces a larger reward margin at every checkpoint, with the gap widening through training (from $+0.005$ at 400 steps to $+0.044$ at 2k steps).

\textbf{SFT.} Both methods reach nearly identical final losses ($1.095$ vs.\ $1.096$), but SDO descends faster in early training ($1.503$ vs.\ $1.526$ at 200 steps). The acceleration occurs where gradient coherence matters most before plateauing.

Across all three paradigms, SDO accelerates convergence without modifying the learning objective. Gains persist through training for DPO and GRPO, while SFT benefits concentrate in the early phase before the loss plateaus. Paired $t$-tests confirm significant mid-training improvements ($p{<}0.05$, Cohen's $d{>}2$); full results are in the Technical Supplement.

\subsection{Per-Cluster Balance across Question Types}

\begin{table}[t]
\centering
\caption{Per-cluster balance on GSM8K/GRPO.``Mean'': average of per-cluster accuracies; B-20\%: bottom-20\% clusters; CV/Gini/Gap: dispersion.}
\label{tab:fairness}
\resizebox{\columnwidth}{!}{%
\begin{tabular}{lccccc}
\toprule
Model & Mean $\uparrow$ & B-20\% $\uparrow$ & CV $\downarrow$ & Gini $\downarrow$ & Gap $\downarrow$ \\
\midrule
GRPO@3k            & 77.02 & 44.94 & 0.2702 & 0.1455 & 100 \\
GRPO+SDO@3k       & 79.72 & 46.77 & 0.2622 & 0.1378 & 100 \\
\midrule
GRPO@4k            & 82.33 & 50.71 & 0.2365 & 0.1216 & 100 \\
GRPO+SDO@4k       & 83.54 & 55.11 & 0.2151 & 0.1121 & 100 \\
\midrule
GRPO@5k            & 83.05 & 55.94 & 0.2106 & 0.1100 & 100 \\
GRPO+SDO@5k       & \textbf{84.21} & \textbf{57.50} & \textbf{0.1955} & \textbf{0.1038} & \textbf{80} \\
\bottomrule
\end{tabular}}
\end{table}

The next question is whether the efficiency gain comes at the expense of uneven data coverage. Table~\ref{tab:fairness} reports per-cluster accuracy on GSM8K, where questions are partitioned into 164 clusters via $k$-means on the same prompt embeddings used by SDO (cluster sizes vary). Results are summarized with five indicators: mean accuracy, bottom-20\% accuracy (B-20\%), coefficient of variation (CV), Gini coefficient, and max--min gap. This is an embedding-space coverage diagnostic rather than a measure of semantic or social fairness.

SDO improves per-cluster balance at every matched step. At 5k steps, B-20\% accuracy rises from 55.94\% to 57.50\%, the Gini coefficient drops from 0.1100 to 0.1038, and the max--min gap falls from 100 to 80 points, with SDO leading on all five indicators. The efficiency gain does not come at the cost of coverage, consistent with Proposition~\ref{prop:exposure}'s coverage preservation guarantee.

\begin{figure}[t]
\centering
\includegraphics[width=1.0\columnwidth]{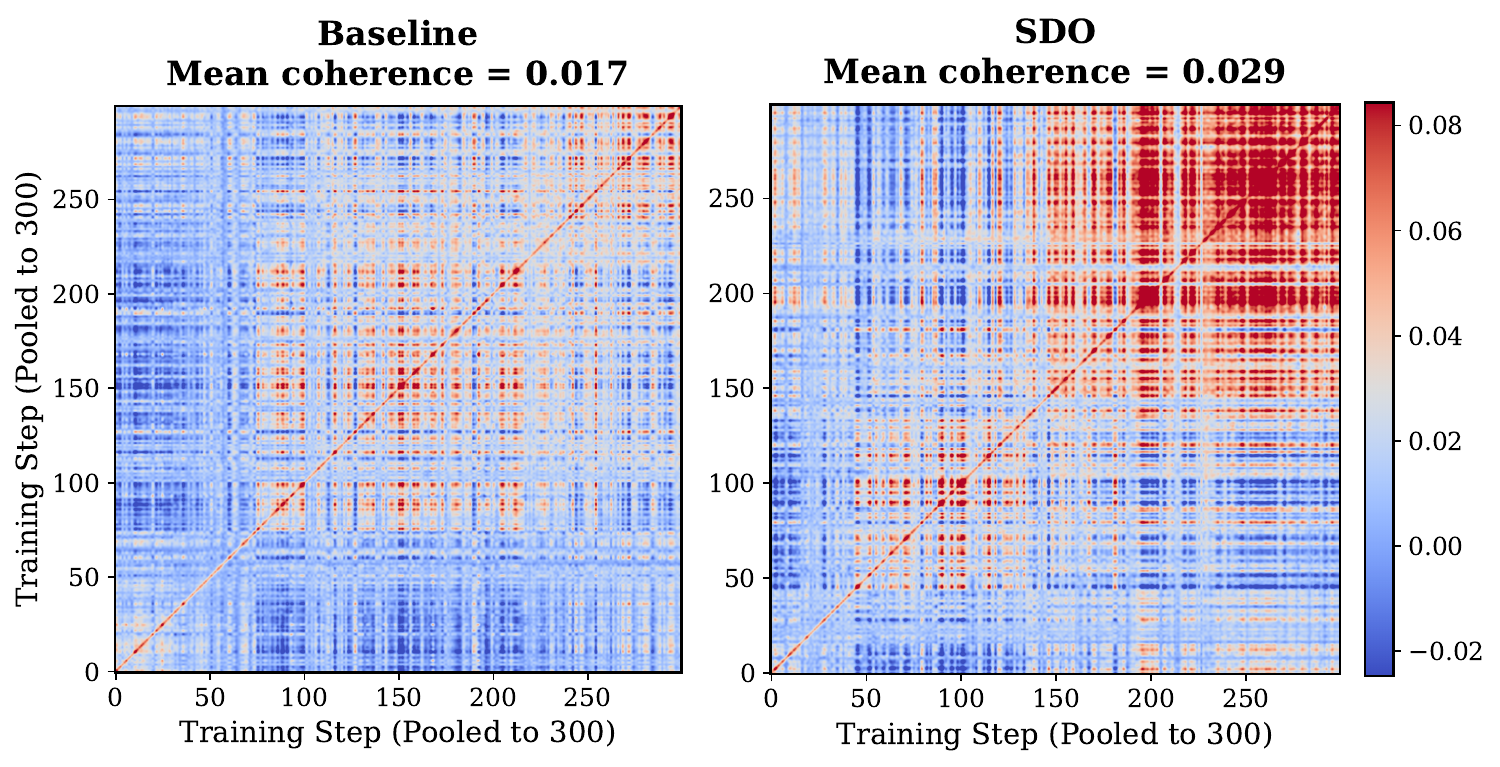}
\caption{Pairwise gradient coherence over the first 3{,}000 GRPO steps. SDO displays broad high-coherence regions (mean 0.29) versus the baseline's sparse pattern (mean 0.17).}
\label{fig:gradient_heatmap}
\end{figure}

\subsection{Gradient Behavior Analysis}
\label{sec:gradient_analysis}

The efficiency gains are consistent with the theoretical mechanism: grouping neighboring samples should produce more coherent gradients. We record gradients of the last trainable layer (as a proxy for gradient behavior) throughout GRPO training under the diagnostic configuration ($r{=}0.1$); these diagnostics provide supportive evidence.

\begin{figure}[t]
\centering
\includegraphics[width=1.0\columnwidth]{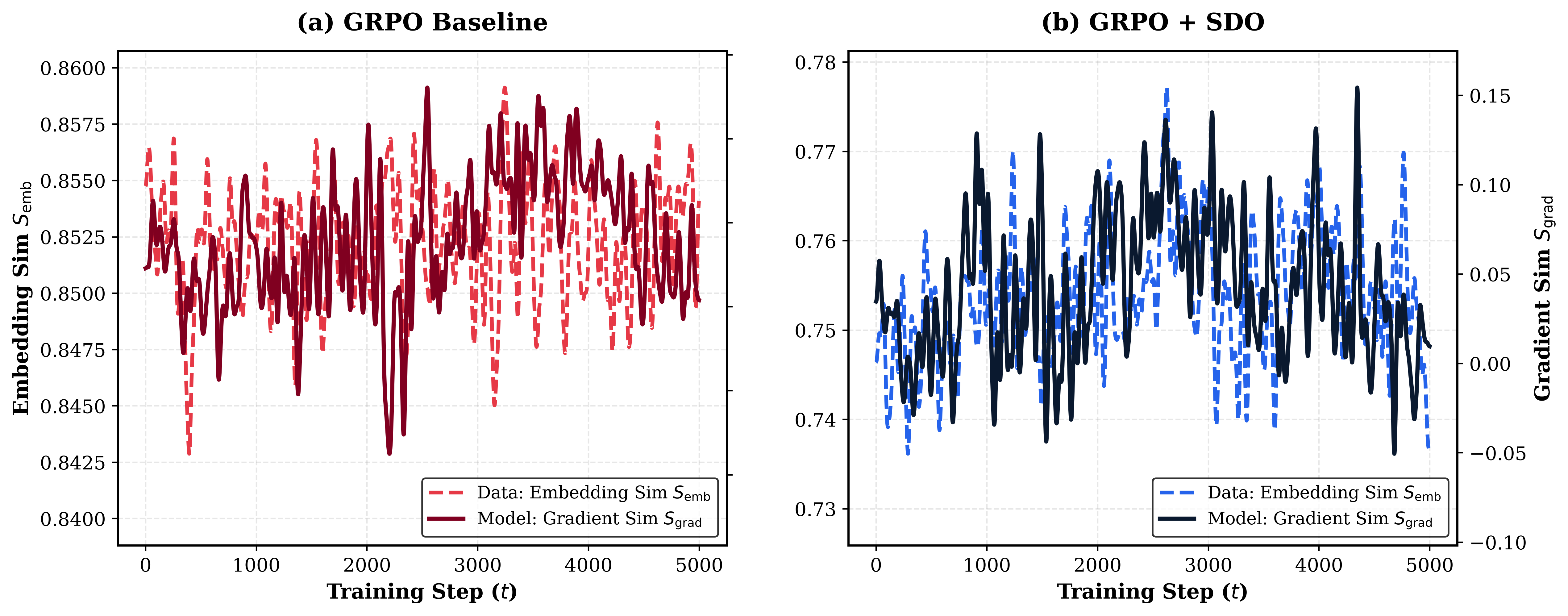}
\caption{Co-evolution of embedding and gradient similarity.}
\label{fig:coevolution}
\end{figure}

Figure~\ref{fig:gradient_heatmap} shows the pairwise gradient coherence matrix over the first 3k steps: the baseline exhibits sparse, scattered coherence (mean 0.17), while SDO displays broad high-coherence regions (mean 0.29), consistent with Theorem~\ref{thm:conflict}'s prediction that locality-aware batching reduces intra-batch gradient conflict. Figure~\ref{fig:coevolution} further tracks whether embedding-space locality induces gradient-space structure: under uniform shuffling, a weak negative correlation is observed (Pearson $\rho{=}-0.09$, $p{<}2{\times}10^{-10}$), while under SDO a significant positive correlation emerges ($\rho{=}0.26$, $p{<}10^{-300}$). These statistics are computed on smoothed, autocorrelated time series and are suggestive rather than conclusive; formal definitions and collection details are in the Technical Supplement.

\subsection{Ablation Studies}
\label{sec:ablation}

All experiments in this section use a single seed (617). Component ablations use $r{=}0.1$; hyperparameter sensitivity (Table~\ref{tab:ablation}) varies all parameters as indicated. Absolute values may differ from the multi-seed means in Table~\ref{tab:main} (which uses the default $r{=}0.2$), but relative trends are consistent.

\begin{table}[t]
\centering
\caption{Hyperparameter sensitivity on GRPO/GSM8K. ``3k''/``5k'': checkpoint accuracy.}
\label{tab:ablation}
\resizebox{\columnwidth}{!}{%
\begin{tabular}{ccccccccc}
\toprule
$K$ & $\Delta\tau$ & $r$ & 3k $\uparrow$ & 5k $\uparrow$ & Mean $\uparrow$ & B-20\% $\uparrow$ & CV $\downarrow$ & Gini $\downarrow$ \\
\midrule
4 & 2 & 0.2 & \textbf{79.68} & 83.32 & 84.21 & 57.50 & 0.1955 & 0.1038 \\
4 & 3 & 0.2 & 79.08 & 82.03 & 83.27 & 56.35 & 0.2094 & 0.1088 \\
8 & 2 & 0.2 & 76.95 & 82.87 & 83.35 & 54.28 & 0.2227 & 0.1123 \\
2 & 2 & 0.2 & 75.28 & 82.18 & 82.87 & 53.83 & 0.2206 & 0.1148 \\
4 & 2 & 0.1 & 79.45 & \textbf{83.92} & \textbf{85.27} & \textbf{61.38} & \textbf{0.1806} & \textbf{0.0952} \\
4 & 1 & 0.2 & 79.00 & 83.55 & 84.43 & 58.95 & 0.1898 & 0.1014 \\
4 & 2 & 0.3 & 79.15 & 83.62 & 84.29 & 57.22 & 0.2072 & 0.1060 \\
\bottomrule
\end{tabular}}
\end{table}

\textbf{Hyperparameter sensitivity.} Table~\ref{tab:ablation} reports checkpoint accuracy (3k, 5k) and per-cluster balance (5k) across hyperparameter settings on GRPO/GSM8K.

\textbf{Effect of $K$.} The neighborhood size $K$ trades off conflict reduction and exploration diversity: larger $K$ improves neighborhood homogeneity and reduces conflict ($\propto\Phi(\theta)$), but decreases distinct compositions per epoch ($\propto N/K$). The 3k checkpoint amplifies the difference: $K{=}4$ reaches 79.68, while $K{=}2$ (75.28) and $K{=}8$ (76.95) lag by 4.4 and 2.7 points respectively, confirming the U-shaped dependence predicted by the trade-off. With $B{=}8$, $K{=}4$ provides the best balance across both $r$ settings.

\textbf{Effect of $\Delta\tau$ and $r$.} $\Delta\tau$ controls how quickly the cold/hot boundary rises: $\Delta\tau{=}1$ (slow) keeps more samples in the fully-retained cold set, yielding the best Mean (84.43) and Gini (0.1014) at $r{=}0.2$; $\Delta\tau{=}3$ (fast) pushes samples into the hot set prematurely, degrading balance (Mean 83.27, CV 0.2094). $\Delta\tau{=}2$ provides a stable middle ground and is adopted as default. The retention ratio $r$ governs pruning strength: $r{=}0.1$ achieves the best per-cluster balance across all metrics (Mean 85.27, CV 0.1806, B-20\% 61.38) by aggressively redirecting the optimizer toward under-explored regions, as Proposition~\ref{prop:exposure}'s negative feedback predicts; $r{=}0.3$ retains too many over-visited samples, diminishing the balancing effect (CV 0.2072); $r{=}0.2$ offers slightly lower but more stable convergence and serves as the main-experiment default.

\begin{table}[t]
\centering
\caption{Component ablation results on GRPO/GSM8K.}
\label{tab:component_ablation}
\resizebox{\columnwidth}{!}{%
\begin{tabular}{lccccc}
\toprule
Variant & 1k & 2k & 3k & 4k & 5k \\
\midrule
Baseline              & 70.43 & 73.46 & 77.18 & 81.50 & 82.34 \\
SDO w/o dynamic KNN  & 69.45 & 71.04 & 74.68 & 80.14 & 82.94 \\
SDO w/o locality     & 69.98 & 73.39 & 78.01 & 81.35 & 82.79 \\
SDO w/o exposure     & \textbf{70.58} & \textbf{74.00} & 78.62 & 82.03 & 83.78 \\
Full SDO             & 70.43 & 73.69 & \textbf{79.45} & \textbf{82.34} & \textbf{83.92} \\
\bottomrule
\end{tabular}}
\end{table}

\begin{table}[t]
\centering
\caption{Per-cluster balance of ablated variants.}
\label{tab:component_fairness}
\resizebox{\columnwidth}{!}{%
\begin{tabular}{lccccc}
\toprule
Variant & Mean $\uparrow$ & B-20\% $\uparrow$ & CV $\downarrow$ & Gini $\downarrow$ & Gap $\downarrow$ \\
\midrule
Baseline              & 83.05 & 55.94 & 0.2106 & 0.1100 & 100 \\
SDO w/o dynamic KNN  & 82.92 & 56.20 & 0.2097 & 0.1116 & 100 \\
SDO w/o locality     & 82.34 & 54.20 & 0.2221 & 0.1149 & 100 \\
SDO w/o exposure     & 82.97 & 53.04 & 0.2282 & 0.1185 & 100 \\
Full SDO             & \textbf{85.27} & \textbf{61.38} & \textbf{0.1806} & \textbf{0.0952} & \textbf{80} \\
\bottomrule
\end{tabular}}
\end{table}

\textbf{Component ablation.} Tables~\ref{tab:component_ablation} (accuracy) and~\ref{tab:component_fairness} (fairness) report three ablated variants on GRPO/GSM8K at 5k step. \textit{SDO w/o dynamic KNN} freezes the neighborhood graph after epoch zero, serving as a proxy for static grouping methods (e.g., $k$-means/HDBSCAN-based batching and EP-Order~\cite{ye2026eporder}). \textit{SDO w/o locality} replaces KNN with uniform shuffling. \textit{SDO w/o exposure} disables pool reconstruction.

\textbf{Effect of locality.} SDO w/o exposure leads all non-full variants at early-to-mid checkpoints (74.00 at 2k, 78.62 at 3k) and achieves competitive final accuracy (83.78), but lags behind full SDO at the critical mid-training point (78.62 vs.\ 79.45 at 3k), indicating that locality provides an early-to-mid convergence advantage that exposure balancing alone cannot fully replicate. SDO w/o locality provides almost no gain over the baseline, confirming that exposure scheduling alone cannot produce coherent gradients.

\textbf{Effect of dynamic KNN.} SDO w/o dynamic KNN, which freezes the neighborhood graph after epoch zero, underperforms the baseline at mid-training (74.68 vs.\ 77.18 at 3k) because its fixed partition becomes misaligned with the evolving pool composition after exposure-driven reconstruction. Though it recovers by 5k (82.94 vs.\ 82.34), the mid-training gap shows that static partitions cannot adapt to the shifting data distribution that exposure balancing creates, justifying the epoch-level KNN reconstruction in Algorithm~\ref{alg:sdo}.

\textbf{Effect of exposure.} Without exposure feedback, SDO w/o exposure concentrates optimization on dense clusters: B-20\% accuracy drops to 53.04 (below the baseline's 55.94), CV rises to 0.228, and the max--min gap remains at 100 points. Full SDO leads all variants on every balance metric (B-20\%=61.38, Gini=0.0952, Gap=80). SDO w/o dynamic KNN underperforms on per-cluster balance because its frozen partition cannot redirect attention across epochs.

\section{Conclusion}

We proposed SDO, a lightweight, plug-and-play data organization framework that forms coherent mini-batches via KNN traversal on frozen embeddings and balances sample exposure across epochs. Without altering any learning objective, SDO accelerates convergence across SFT, DPO, and GRPO, most consistently in the early-to-mid phase, while producing more coherent gradients and more balanced per-cluster accuracy. A theoretical analysis links representation-space locality to reduced intra-batch gradient conflict, supported by co-evolution diagnostics. These results suggest that exposure-driven data organization is a practical and complementary lever for efficient post-training.

\bibliography{aaai2027}


\end{document}